\pdfoutput=1

\documentclass[11pt]{article}

\usepackage{EMNLP2023}

\usepackage{times}
\usepackage{latexsym}

\usepackage[T1]{fontenc}

\usepackage[utf8]{inputenc}

\usepackage{microtype}

\usepackage{inconsolata}

\let\origfootnote\footnote
\renewcommand{\footnote}[1]{\kern.1em\origfootnote{#1}}
\newcommand{\punctfootnote}[1]{\kern-.1em\origfootnote{#1}}
\usepackage{booktabs}

\usepackage{tikz}
\usetikzlibrary{positioning}
\tikzset{every node/.style={draw, minimum width=2cm, minimum height=1cm, font={\small}}}
\tikzset{->,>={latex}}
\newcommand{\txtbox}[1]{\begin{minipage}{2cm}\centering #1\end{minipage}}

\title{CUNI Submission to MRL 2023 Shared Task on Multi-lingual Multi-task Information Retrieval}

\author{Jind\v rich Helcl\footnotemark[1] \and Jind\v rich Libovick\'{y}\footnotemark[1] \\
  Charles University, Faculty of Mathematics and Physics \\
  Institute of Formal and Applied Linguistics \\
  V Hole\v{s}ovi\v{c}k\'{a}ch 747/2, 180 00 Prague, Czech Republic \\
  {\tt \{helcl,libovicky\}@ufal.mff.cuni.cz} \\}

\begin{document}
\maketitle
\begin{abstract}
\let\thefootnote\relax\footnotetext{$^\ast$ The author order was determined by a coin toss.}
%
We present the Charles University system for the MRL~2023 Shared Task on Multi-lingual Multi-task Information Retrieval.
The goal of the shared task was to develop systems for named entity recognition and question answering in several under-represented languages. 
Our solutions to both subtasks rely on the translate-test approach. 
We first translate the unlabeled examples into English using a multilingual machine translation model.
Then, we run inference on the translated data using a strong task-specific model.
Finally, we project the labeled data back into the original language.
To keep the inferred tags on the correct positions in the original language, we propose a method based on scoring the candidate positions using a label-sensitive translation model.
In both settings, we experiment with finetuning the classification models on the translated data.
However, due to a domain mismatch between the development data and the shared task validation and test sets, the finetuned models could not outperform our baselines.
\end{abstract}

\section{Introduction}

Pre-trained language models reach state-of-the-art results in most current natural language processing (NLP) tasks. Whereas in high-resource languages such as English, we observe in-context learning capabilities and emergent abilities \citep{wei2022emergent}, in less-resourced languages, the results are more modest \citep{lai2023chatgpt}, mostly due to the lack of necessary data needed to train really large models. Moreover, there is usually not enough task-specific data available in these languages.
This leads to attempts to reuse the (high-resource) language model capabilities in other (low-resource) languages.

Most of the proposed methods are either based on transfer learning \citep{lauscher-etal-2020-zero,yu-joty-2021-effective,zheng-etal-2021-consistency,schmidt-etal-2022-dont} or machine translation (MT), both during training and at test time (e.g. mentioned as a baseline by \citealp{conneau-etal-2020-unsupervised,conneau-etal-2018-xnli}).

The MRL~2023 Shared Task on Multi-lingual Multi-task Information Retrieval aims to explore these methods further, applied to many low-resource languages. The participants
were tasked to build models for two subtasks: named entity recognition (NER) and question answering (QA).

The shared task setup is inspired by the XTREME-UP dataset \citep{ruder2023xtreme}, which focuses on the most needed tasks for under-resourced languages: gathering data in a digital form (speech recognition, optical character recognition, transliteration) and making information in these languages accessible (NER, QA, retrieval for QA).
This dataset contains a relatively small amount of data for multiple tasks on low-resource languages, featuring 88 languages in total, including QA datasets for 4 languages and NER datasets for 20 languages.

The shared task evaluation campaign focused on Igbo, Indonesian (QA only), Alsatian,\punctfootnote{Mistakenly labeled as Swiss German on the task website.} Turkish, Uzbek (QA only), and Yoruba. Out of these languages, only Indonesian is among the XTREME-UP QA datasets, and only Igbo and Yoruba have available NER task data in the benchmark. Upon releasing the validation data close to the end of the campaign, Azerbaijani was added as a surprise language for evaluation (with no data for QA or NER in XTREME-UP).

This setting left the participants with a choice to either collect external training data for languages not present in the benchmark (which was implicitly discouraged by the inclusion of the surprise language) or to develop language-agnostic systems.

Even though a lot of research effort is invested in developing systems that are inherently multilingual, typically based on pre-trained massively multilingual models \citep[inter alia]{artetxe-schwenk-2019-massively,lauscher-etal-2020-zero,pfeiffer-etal-2020-mad,xue-etal-2021-mt5}, our submission is based on the translate-test approach that was recently shown to perform better than the community previously thought \citep{artetxe2023revisiting}.
We rely on the translation quality of a multi-lingual machine translation (MT) system, combined with the
strong performance of pre-trained LLMs in English. The main ideas that are common to our approaches to both subtasks are described in Section \ref{sec:common}. The particularities of our models which are specific to the NER and QA subtasks, are presented in Sections \ref{sec:ner} and \ref{sec:qa}, respectively, including our results on those tasks.

Overall, we find that the translate-test approach can be useful in a multilingual setting. Our results do not outperform supervised, language-specific models, but are considerably better than zero-shot approaches.

To maximize reproducibility, we built our systems using an automated end-to-end development pipeline implemented in Snakemake \citep{snakemake}; we release the code online.\punctfootnote{\url{https://github.com/ufal/mrl2023-multilingual-ir-shared-task}}

\section{Main Ideas}
\label{sec:common}

In both tasks, we employ the translate-test approach,
which can be summarized in the following three steps:
First, we translate unlabeled examples from the task language into English using a multilingual MT model.
Second, we use a pre-trained LLM to perform the task which assigns the labels to the example.
Third, we use a label-aware translation model to project the inferred labels back to the target language.

\paragraph{Translation into English.}
In the first step, we translate the unlabeled data into English.
In both subtasks, we use the NLLB-3.3B\footnote{\url{https://huggingface.co/facebook/nllb-200-3.3B-easyproject}} multilingual MT model \citep{costa2022no}. 
We discuss the task-specific data processing details further in Sections \ref{sec:ner} and \ref{sec:qa}.

\paragraph{Task-specific models.}
In each subtask, we apply a RoBERTa-large model, which has been finetuned on the task \citep{liu2019roberta}. This predicts labels for the English 
data. For NER, these are BIO-encoded labels, marking the span and type of each named entity in the example. Specifically, the output is a sequence of labels of the same length as the input sentence. For QA, the labels mark a span in the context representing the answer. This is encoded using two numbers, which denote character offsets in the detokenized version of the context paragraph.

\paragraph{Translation into the target language.}

The translate-test approach is less challenging when the labels are language-independent, which is also the case of both subtasks. However, span labeling tasks (such as NER and QA) require careful handling of the projection of the spans, i.e., we need to find the corresponding spans in the original language.

Our systems adopt the label projection method for cross-lingual transfer, originally meant for the translate-train approach \citep{chen-etal-2023-frustratingly}. The authors of the paper finetune the NLLB model\footnote{\url{https://huggingface.co/ychenNLP/nllb-200-3.3B-easyproject}}
 to translate texts containing inserted tags so that the tags generated in the translation mark equivalent parts of the source sentence. In contrast to the original use-case of generating the whole target sentence with tags, we already know the target sentence in the shared task scenario. Therefore, we are only interested in the placement of the tags.

To find the best possible placement of the tags, we propose to use the aforementioned finetuned model as a scorer. We place the tags at all possible positions (subject to minimum/maximum span length constraints) and select the highest-scoring candidate. We then either reconstruct the label sequence (in the case of NER) or extract the appropriate passage from the context (for QA).

\section{Named Entity Recognition}
\label{sec:ner}

The goal of the NER subtask was to classify words and phrases into one of four categories: 
person (PER), organization (ORG), location (LOC), and date (DAT). Since most state-of-the-art NER classifiers (including the one we used) use a richer set of labels, we apply rule-based mapping to reduce the label set to the four categories: geopolitical entities (GPE)  and facilities (FAC) are replaced with LOC, time with DAT.

The XTREME-UP benchmark contains two NER datasets, MasakhaNER \citep{adelani-etal-2021-masakhaner} and MasakhaNER~2 \citep{adelani-etal-2022-masakhaner}, both using texts from local news stories and covering 10 and 20 African languages respectively.

\begin{figure*}
\centering
\begin{tikzpicture}

\node (src) {\txtbox{Sentence in the source language}};
\node (mt) [right = 15pt of src] {\txtbox{MT into English}};
\draw (src) -- (mt);

\node (tok) [right = 15pt of mt] {Tokenizer};
\draw (mt) -- (tok);

\node (ner) [right = 15pt of tok] {\txtbox{NER in English}};
\draw (tok) -- (ner);

\node (project) [right = 15pt of ner] {\begin{minipage}{2cm}\centering Project NER spans back to source \end{minipage}};
\draw (ner) -- (project);
\draw (src.north) to[out=10,in=170] (project.north);

\draw (project) -- ++(40pt, 0pt);
\end{tikzpicture}
\caption{A scheme of the NER translate-test pipeline.}%
\label{fig:ner}
\end{figure*}
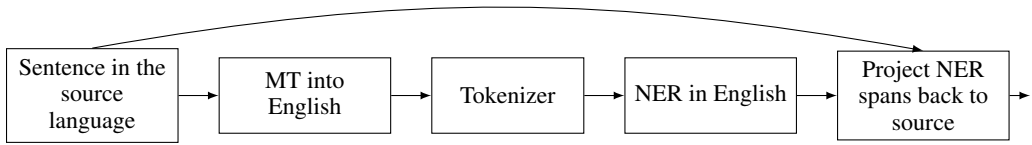

The scheme of the translate-test pipeline for this task is shown in Figure~\ref{fig:ner}.

\paragraph{English NER models.} 
For experiments with English NER, we use the tNER toolkit \citep{ushio-camacho-collados-2021-ner},
which provides several models for this task. We selected two RoBERTa-based models for experiments, finetuned either on Ontonotes5 \citep{hovy-etal-2006-ontonotes} or on the CoNLL2003 \citep{tjong-kim-sang-de-meulder-2003-introduction} dataset. 
We found that using the model built with Ontonotes5\footnote{\url{https://huggingface.co/tner/roberta-large-ontonotes5}} leads to better results than the one finetuned on the CoNLL data.\punctfootnote{\url{https://huggingface.co/tner/roberta-large-conll2003}}
Ontonotes5 contains news stories, television and radio transcripts, and web pages. The CoNLL 2003 data is international news from 1996--1997. This means there is a domain mismatch between the most available named entity recognizers and the MasakhaNER datasets.

\paragraph{Finetuning.} To overcome the domain mismatch, we finetuned the tNER models using the MasakhaNER data. We translated the MasakhaNER training data into English and performed the span projection the same way as at inference time. The finetuning step serves not only as domain adaptation to news stories from the non-English speaking world but also as an adaptation to texts which have been automatically translated from low-resource languages.

\paragraph{Results.}
Table \ref{tab:ner_masakhaner1} presents the results on the MasakhaNER~1 dataset. Our translate-test approach significantly outperforms zero-shot transfer from English using the XLM-R \citep{conneau-etal-2020-unsupervised} and XLM-V \citep{liang2023xlmv} models; however, there is still a large performance gap between the translate-test approach and supervised in-language training.

\begin{table*}
    \centering\footnotesize
    \begin{tabular}{l cccccccccc c} \toprule
 & amh & hau & ibo & kin & lug & luo & pcm & swa &  wol & yor & AVG \\ \midrule
Best baseline (supervised) & 78.0 & 91.5 & 87.7 & 77.8 & 84.7 & 75.3 & 90.0 & 89.5 & 86.3 & 83.7 & 84.4 \\
XLM-R (zero-shot) & 25.1 & 43.5 & 11.6 & 9.4 & 9.5 & 8.4 & 36.8 & 48.9 & 5.3 & 10.0 & 20.9 \\
XLM-V (zero-shot) & 20.6 & 35.9 & 45.9 & 25.0 & 48.7 & 10.4 & 38.2 & 44.0 & 16.7 & 35.8 & 32.1 \\ \midrule
Spacy & 59.2 & 58.3 & 57.7 & 48.5 & 52.6 & 45.8 & 9.0 & 60.0 & 48.1 & 47.2 & 48.6 \\
tNER: ConLL2003 & 57.3 & 66.7 & 72.8 & 57.0 & 69.4 & 49.7 & 65.8 & 69.4 & 53.7 & 59.3 & 62.1 \\
tNER: Ontonotes5 & 60.8 & 62.8 & 73.3 & 60.2 & 69.9 & 52.5 & 74.2 & 70.1 & 51.4 & 57.5 & 63.3 \\
+ finetuning& 61.8 & 70.0 & 76.4 & 65.4 & 70.2 & 57.5 & 77.9 & 74.5 & 58.2 & 59.6 & 67.2 \\ \bottomrule
\end{tabular}
    \caption{F1 scores on the MasakhaNER 1 dataset.}%
    \label{tab:ner_masakhaner1}
\end{table*}

The MasakhaNER~2 results are shown in Table~\ref{tab:ner_masakhaner2}.  Similarly to MasakhaNER~1, our results are strictly worse than supervised training. The second line of the table shows the results of a model trained on MasakhaNER~1 but tested on MasakhaNER~2, which contains ten more languages than the first dataset. The results on these additional languages (shown in boxes) mark zero-shot transfer between African languages. Our translate-test approach via English is better than zero-shot using African languages for 5 of the 10 languages.

\begin{table*}
\centering\footnotesize
\resizebox{\textwidth}{!}{\begin{tabular}{l cccccccccc c}
\toprule
 & bam & bbj & ewe & fon & hau & ibo & kin & lug & mos & nya \\ \midrule
Best supervised in paper & 82.2 & 75.2 & 90.3 & 82.7 & 87.4 & 89.6 & 87.5 & 89.6 & 76.4 & 92.4 \\
Trained on MasakhaNER 1 & \fbox{50.9} & \fbox{49.8} & \fbox{76.2} & \fbox{57.1} & 88.7 & 90.1 & 87.6 & 90.0 & \fbox{75.0} & \fbox{80.4} \\ \midrule
Spacy & 38.1 & 16.8 & 57.0 & 39.9 & 48.1 & 52.0 & 55.3 & 65.7 & 31.5 & 53.0 \\
tNER: ConLL2003 & 49.0 & 21.9 & 67.5 & 51.7 & 66.3 & 64.9 & 60.6 & 74.8 & 42.5 & 66.0\\
tNER: Ontonotes5 & 46.8 & 20.5 & 67.3 & 48.8 & 63.6 & 63.6 & 64.6 & 75.2 & 39.8 & 69.0 \\
+ finetuning & \bf 60.9 & 25.9 & 73.7 & 53.0 & 67.0 & 75.3 & 65.7 & 75.2 & 44.7 & 72.1 \\

\midrule
& pcm & sna & swa & tsn & twi & wol & xho & yor & zul & AVG \\ \midrule

Best supervised in paper & 90.1 & 96.2 & 92.7 & 89.4 & 81.8 & 86.8 & 89.9 & 89.3 & 90.6 & 87.1 \\
Trained on MasakhaNER 1 & 90.2 & \fbox{42.5} & 93.1 & 79.4 & \fbox{57.3} & 87.0 & \fbox{47.4} & 89.7 & \fbox{64.3} & 74.0 \\
\midrule
Spacy & 52.5 & 60.6 & 67.4 & 63.4 & 53.7 & 46.5 & 47.7 & 42.3 & 56.2 & 49.9 \\
tNER: ConLL2003  & 67.7 & 69.7 & 70.8 & 74.8 & 67.6 & 61.9 & 67.0 & 52.9 & 66.0 & 61.2 \\
tNER: Ontonotes5 & 72.8 & 72.4 & 72.7 & 73.1 & 62.0 & 57.7 & 67.9 & 55.6 & 70.3 & 61.3 \\
+ finetuning & 79.2 & \bf 81.7 & 75.1 & 76.2 & \bf 68.4 & 65.6 & \bf 75.8 & 60.5 & \bf 70.4 & 66.7 \\

\bottomrule
\end{tabular}}
\caption{F1 scores on the MasakhaNER 2 dataset. The numbers in boxes denote zero-shot transfer between African languages (i.e., languages that are in MasakhaNER~2 but not in MasakhaNER~1). Bold numbers are results where our approach is better than the zero-shot transfer between African languages.}%
\label{tab:ner_masakhaner2}
\end{table*}

When compared to related work, our results (average score 61.3\%) without finetuning outperform transfer from English using mDeBERTav3 \citep{he2023deberta} (average score 55.5\%). However, they are worse when compared to the translate-train results reported by \citet{chen-etal-2023-frustratingly} (average score 63.4\%) that used additional parallel data with projected labels for training.


On both MasakhaNER benchmarks, the Ontonotes5 model is slightly better than CoNLL 2003. Finetuning (which involves training data of the respective datasets) leads to consistent improvements. On MasakhaNER~2, the finetuned model outperforms \citet{chen-etal-2023-frustratingly}; however, the training data setups are not easily comparable.

The results on the shared task validation data are in Table~\ref{tab:ner_valid}. 
Because of the domain mismatch (the shared task validation data are not local news but rather Wikipedia articles), the original Ontonotes5 model performs better. Based on this observation, we decided to use the pipeline using the original Ontonotes5 model \emph{without} finetuning.
We omit Yoruba from calculating the average score because most entities are left without annotation in the data.

\begin{table}
\centering\footnotesize
\begin{tabular}{l cccc c}
\toprule
               & als  & aze  & tur  & yor & AVG \\ \midrule
CoNLL 2003     & 38.4 & 54.6 & 48.0 & 4.4 & 47.0 \\
Ontonotes5     & 40.3 & 62.8 & 54.7 & 3.1 & 52.4 \\
+ finetuned    & 40.9 & 62.3 & 51.3 & 5.9 & 51.7 \\ \bottomrule
\end{tabular}
\caption{Results on the shared task validation data. The average does not include Yoruba.}%
\label{tab:ner_valid}
\end{table}

\section{Question Answering}
\label{sec:qa}

\begin{figure*}
\centering
\begin{tikzpicture}

\node[minimum height=20pt] (src_q) {\txtbox{Question and}};
\node (src_c) [below=0pt of src_q] {\txtbox{Context in the source language}};

\node (mt_q) [above right = -10pt and 50pt of src_q] {\txtbox{MT into English}};
\draw (src_q) -- (mt_q);

\node (split_c) [below right = -10pt and 15pt of src_c] {\txtbox{Sentence split}};
\draw (src_c) -- (split_c);
\node (mt_c) [right = 10pt of split_c] {\txtbox{MT into English}};
\draw (split_c) -- (mt_c);

\node (class) [right=65pt of mt_q] {\txtbox{No-answer classifier}};
\draw (mt_q) -- (class);
\draw (mt_c) -- (class);

\node (qa) [right= 15pt of mt_c] {\txtbox{QA in English}};
\draw (mt_q) -- (qa);
\draw (mt_c) -- (qa);

\node (project) [right = 15pt of qa] {\txtbox{Project the answer span back}};
\draw (qa) -- (project);
\draw (split_c.south) to[out=-10,in=190] (project.south);

\node (merge) [above right = -5pt and 15pt of project] {Merge};
\draw (project) -- (merge);
\draw (class) -- (merge);
\draw (merge) -- ++(40pt, 0pt);

\end{tikzpicture}
\caption{A scheme of the QA translate-test pipeline.}%
\label{fig:qa}
\end{figure*}
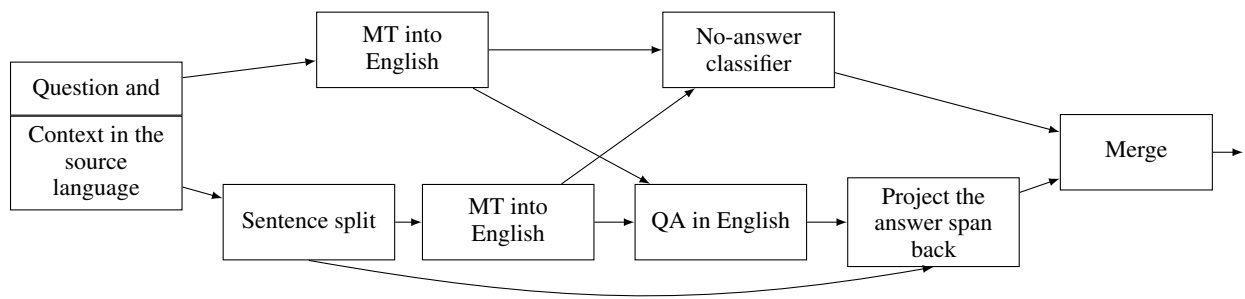

The goal of this task is to find an answer to a given question within a given context.
In the generative version of this task, the answer may not be taken from the context directly.
Figure \ref{fig:qa} shows the question-answering processing pipeline we use in our experiments.

\paragraph{Data Preprocessing.}
The XTREME-UP datasets for QA consist of three fields: The question, the context, and the answer.
Since the context might be several sentences long, we apply sentence splitting using \texttt{wtpsplit} \citep{minixhofer-etal-2023-wheres}. Since the toolkit does not support Alsatian or Swahili, we use the English variant instead. After sentence splitting, we translate everything into English using the NLLB model.
Since NLLB does not support Alsatian, we set the source language to German.

\paragraph{Answering Questions.}
For the extractive question answering task, we use a RoBERTa-based model\footnote{\url{https://huggingface.co/deepset/roberta-large-squad2}} finetuned for question answering to mark the
answer spans in the English context. Once the spans are found, we insert tags into the English sentence.
To find the right spans in the original language, we use the tag-preserving NLLB model as a scorer
and select the highest-scoring span according to the model.

\paragraph{No Answer Classification.}
Since there are examples with no answer in XTREME-UP, we train a classifier to detect such cases. We again use the QA-tuned RoBERTa, which we finetune on 3 epochs of the translated XTREME-UP data.
We set the learning rate to $10^{-5}$, weight decay to 0.01, and keep the default values of the rest of the hyper-parameters. The classifier achieves 93\% accuracy on the development set. However, because the shared task validation set contains only a very small amount of examples with no answer, we decided not to use this classifier in our submissions.

\paragraph{In-domain Finetuning.}
We also implemented in-domain finetuning of the QA RoBERTa model on the XTREME-UP dataset translated into English. Because the answers are represented as spans within the context, we use the same technique to project the spans onto the English translation of the context as we use in span projecting to the original language.
Performing grid search and measuring model performance on the development set, we found a learning rate of $5\times10^{-6}$, gradient norm of 1, warmup ratio of 0.5, and weight decay of 0.1 are the most suitable hyper-parameters.

\paragraph{Using Generative Models.}
We noticed that the shared task validation data did not actually contain examples of extractive 
question answering. Instead, the answers were likely written by a human annotator. Therefore, we decided
also to submit a contrastive experiment using a generative model, namely Llama 2 \citep{touvron2023llama}.\punctfootnote{\url{https://huggingface.co/meta-llama/Llama-2-13b-chat-hf}} For the generation, we use the prompt \texttt{"Context: \{context\} Question: \{question\} Short answer:"}. We apply rule-based post-processing to remove potential continuations generated after the answer. Details can be found in the corresponding \texttt{Snakefile} in the code repository.  

\begin{table*}
\centering\footnotesize
\begin{tabular}{l rrrrrr r }
\toprule
   & als  & aze  & ind & tur & uzn  & yor & AVG \\ 
\midrule
Roberta-large & 8.34	& 17.87	& 30.36	& 14.84	& 23.13	& 18.56	& 18.85 \\
Finetuned & 7.88	& 15.83	& 30.94	& 12.64	& 18.60	& 19.65	& 17.59 \\
\midrule
Llama 2 & 17.43&31.96&34.61&24.51&30.81&19.80&26.52 \\
\bottomrule
\end{tabular}
\caption{Question answering results on the shared task validation data (chrF).}%
\label{tab:qa_valid}
\end{table*}

\paragraph{Results.}
Table \ref{tab:qa_valid} shows the results of the shared task validation set. Since there is a considerable domain mismatch between the XTREME-UP dataset and the shared task validation and test sets,
we see that the in-domain finetuning does not improve the performance -- we, therefore, use the baseline systems as our primary submission. Using the generative model, however, achieves a substantial improvement. Because the task was originally aimed at extractive QA, we decided to submit the generative model as a contrastive experiment.

\section{Conclusions and Discussion}

The research community long overlooked the translate-test approach until recently, when \citet{artetxe2023revisiting} showed that it might outperform both translate-train and cross-lingual transfer with sufficiently strong machine translation systems.

With the increasing number of attempts to use large generative language models in cross-lingual setups, we speculate that the translate-test approach will become an important baseline that might not be easy to cross. Methods that work well with multilingual encoders enforce alignment of the intermediate representation \citep[inter alia]{wu-dredze-2020-explicit,hammerl-etal-2022-combining,pfeiffer-etal-2022-lifting}. However, in generative setups, this would lead to undesirable language mixing \citep{li-murray-2023-zero}. Generative models are also known not to be consistent across languages \citep{lai2023chat,wang2023seaeval}. Translate-test does not suffer from either of these disadvantages.

We successfully tested the translate-test method in the shared task setup involving span-labeling tasks. We translated the input into English, performed the task using state-of-the-art English models, and projected the results back to the original language. The main technical challenge is that after labeling the spans in English, we need to find the corresponding span in the original text. For that purpose, we used an MT model specifically finetuned to preserve tags encoded as brackets. Furthermore, we finetuned the task-specific models on XTREME-UP data automatically translated into English.

Although the shared task claimed to be based on the XTREME-UP benchmark, the actual shared task data have many different characteristics. Instead of local news outlets, the NER data used Wikipedia text, often on generic topics rather than local ones. The QA validation and test data were abstractive, not extractive. Because of that, our finetuned models performed worse than the original ones. Also, generative QA using LlaMA 2 outperformed our original extractive system.

The final results show that building a translate-test pipeline is a viable approach to both cross-lingual NER and QA.

\section*{Limitations}

Both validation and test datasets from the shared task are considerably small, especially for QA, where they contain only around 100 examples per language. This might lead to an unreliable comparison between the submitted systems.

The paper does not contain experimental results that would sufficiently back stronger claims about translate-test approaches. We made decisions that appeared to lead to a good performance in the context of the shared task. However, the paper misses ablations that would reliably show that the span projection method is the best. More importantly, this paper does not compare our results with a strong system based on cross-lingual transfer.

None of the system authors speak the languages in the shared task, and neither is particularly familiar with the culture of the respective language communities. The authors did not check the system 
outputs for harmful or otherwise inappropriate content.

\section*{Acknowledgments}

The work was supported by the Charles University project PRIMUS/23/SCI/023. 
The work described herein has been using services provided by the LINDAT/CLARIAH-CZ Research Infrastructure (https://lindat.cz), supported by the Ministry of Education, Youth and Sports of the Czech Republic (Project No. LM2023062).

\bibliography{anthology,custom}
\bibliographystyle{acl_natbib}

\end{document}